# New Advances in Inference by Recursive Conditioning

**David Allen** and **Adnan Darwiche**
Computer Science Department
University of California
Los Angeles, CA 90095
{dlallen,darwiche}@cs.ucla.edu

## Abstract

Recursive Conditioning (RC) was introduced recently as an any–space algorithm for inference in Bayesian networks which can trade time for space by varying the size of its cache at the increment needed to store a floating point number. Under full caching, RC has an asymptotic time and space complexity which is comparable to mainstream algorithms based on variable elimination and clustering (exponential in the network treewidth and linear in its size). We show two main results about RC in this paper. First, we show that its actual space requirements under full caching are much more modest than those needed by mainstream methods and study the implications of this finding. Second, we show that RC can effectively deal with determinism in Bayesian networks by employing standard logical techniques, such as unit resolution, allowing a significant reduction in its time requirements in certain cases. We illustrate our results using a number of benchmark networks, including the very challenging ones that arise in genetic linkage analysis.

## 1    INTRODUCTION

Recursive Conditioning, RC, was recently proposed as an any–space algorithm for exact inference in Bayesian networks [3]. The algorithm works by using conditioning to decompose a network into smaller subnetworks that are then solved independently and recursively using RC. It turns out that many of the subnetworks generated by this decomposition process need to be solved multiple times redundantly, allowing the results to be stored in a cache after the first computation and then subsequently fetched during further computations. This gives the algorithm its any–space behavior since any number of results may be cached, where a "result" corresponds to a floating point number which represents a probability.

The ability of RC to provide a refined framework for time–space tradeoff has been explored in some depth recently [1] and will not be the subject of this paper. Instead, we focus on two key aspects of RC. First, its actual space requirements under full caching, as compared to the space requirements needed by variable elimination (VE) and jointree (JT) algorithms [4; 21; 11; 3; 10], where we show experimentally, and argue theoretically, for the modest space demands of RC as compared to other methods. Second, we discuss the ability of RC in taking advantage of some very effective techniques, which appear to be best realized in a conditioning setting. Among these techniques are the ability of RC to accommodate any representation of network parameters (CPTs) without requiring an algorithmic change, and the ability of RC to exploit network determinism by easily incorporating standard techniques from the SAT community, including unit resolution.

## 2    RECURSIVE CONDITIONING

RC works by using conditioning and case analysis to decompose a network into smaller subnetworks that are solved independently and recursively. The algorithm is driven by a structure known as a decomposition tree (dtree), which controls the decomposition process at each level of the recursion. We will first review the dtree structure and then discuss RC.

### 2.1    DTREES

**Definition 1** [3] A _dtree_ for a Bayesian network is a full binary tree, the leaves of which correspond to network variables. If a leaf node $t$ corresponds to variable $X$ with parents $\mathbf{U}$, then vars$(t)$ is defined as $\{X\} \cup \mathbf{U}$.

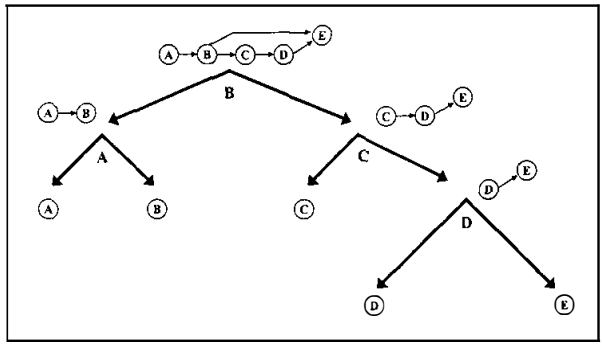

Figure 1: An example dtree.

Dtrees correspond to *branch-decompositions* as known in the graph–theoretic literature [19]. In particular, dtrees are for branch–decompositions what jointrees are for tree–decompositions [18].

Figure 1 depicts a simple dtree. The root node $t$ of the dtree represents the entire network. To decompose this network, the dtree instructs us to condition on variable $B$, called the cutset of root node $t$. Conditioning on a set of variables leads to removing edges outgoing from these variables, which for a cutset is guaranteed to disconnect the network into two subnetworks, one corresponding to the left child of node $t$ and another corresponding to the right child of node $t$; see Figure 1. This decomposition process continues until a boundary condition is reached, which is a subnetwork that has a single variable.

We will now present some notation needed to define additional concepts with regard to a dtree. The notation $t_l$ and $t_r$ will be used for the left child and right child of node $t$, and the function vars will be extended to internal nodes $t$: $\text{vars}(t) \stackrel{def}{=} \text{vars}(t_l) \cup \text{vars}(t_r)$. Each node in a dtree has three more sets of variables associated with it. The first two of these sets are used by the RC algorithm, while the third set is used to analyze the complexity of the algorithm.

**Definition 2** *The <u>cutset</u> of internal node $t$ in a dtree is:* $\text{cutset}(t) \stackrel{def}{=} \text{vars}(t_l) \cap \text{vars}(t_r) - \text{acutset}(t)$, *where* $\text{acutset}(t)$ *is the union of cutsets associated with ancestors of node $t$ in the dtree. The <u>context</u> of node $t$ in a dtree is:* $\text{context}(t) \stackrel{def}{=} \text{vars}(t) \cap \text{acutset}(t)$. *The <u>cluster</u> of node $t$ in a dtree is:* $\text{cutset}(t) \cup \text{context}(t)$ *if $t$ is a non-leaf, and as $\text{vars}(t)$ if $t$ is a leaf.*

The *width* of a dtree is the size of its largest cluster $-1$. It is known that the minimum width over all dtrees for a network is the *treewidth* of corresponding network [3]. The *context width* of a dtree is the size of its largest context. Moreover, the minimum context width over all dtrees is the *branchwidth* of corresponding network

**Algorithm 1** RC($t$): Returns the probability of evidence **e** recorded on the dtree rooted at $t$.

1: If $t$ is a leaf node, return LOOKUP($t$)
2: $\mathbf{y} \leftarrow$ recorded instantiation of $\text{context}(t)$
3: If $\text{cache}_t[\mathbf{y}] \neq$ nil, return $\text{cache}_t[\mathbf{y}]$
4: $p \leftarrow 0$
5: **for** instantiations **c** of uninst. vars in $\text{cutset}(t)$ **do**
6:    record instantiation **c**
7:    $p \leftarrow p + RC(t_l)RC(t_r)$
8:    un–record instantiation **c**
9:    $\text{cache}_t[\mathbf{y}] \leftarrow p$
10: return $p$

**Algorithm 2** LOOKUP($t$)

$X \leftarrow$ variable associated with $t$
**if** $X$ is instantiated **then**
   $x \leftarrow$ recorded instantiation of $X$
   $\mathbf{u} \leftarrow$ recorded instantiation of $X$'s parents
   return $\text{Pr}(x|\mathbf{u})$
**else**
   return 1

[19]. It is also known that the treewidth is no more than a constant factor from branchwidth [19].

The *cutset* of a dtree node $t$ is used to decompose the network associated with node $t$ into the smaller networks associated with the children of $t$. That is, by conditioning on variables in $\text{cutset}(t)$, one is guaranteed to disconnect the network associated with node $t$. The *context* of dtree node $t$ is used to cache results: Any two computations on the subnetwork associated with node $t$ will yield the same result if these computations occur under the same instantiation of variables in $\text{context}(t)$. Hence, a cache is associated with each internal dtree node $t$, which stores the results of such computations (probabilities) indexed by instantiations of $\text{context}(t)$. This means that the size of a cache associated with dtree node $t$ can grow as large as the number of instantiations of $\text{context}(t)$. The total space requirement of RC is then $\sum_t \|\text{context}(t)\|$, where $\|S\|$ represents the number of instantiations of variables $S$.

## 2.2 INFERENCE USING RC

Given a Bayesian network and a corresponding dtree with root $t$, the RC algorithm given in Algorithm 1 can be used to compute the probability of evidence **e** by first "recording" the instantiation **e** and then calling $RC(t)$, which returns the probability of **e**.

Note that Line 9 is where space is used by RC as it is on this line that a cache entry is filled. When every computation is cached, RC uses $O(n \exp(w))$ time, where $n$ is the number of nodes in the network and $w$ is the width of used dtree. This corresponds to the complexity of JT algorithms, assuming that the dtree is generated from a jointree, and to VE algorithms, assuming that the dtree is generated from an elimination order [3]. As for space requirements under full caching, RC uses $O(n \exp(w_c))$ space where $w_c$ is the context width of used dtree ($w_c \leq w + 1$).[1]

Table 1: Space/Time Requirements for Various Formalisms. The time for Hugin is broken down into the time to compile a jointree (excluding triangulation), and the time to propagate. The experiments used a Pentium 4, 2.4 Ghz processor, with 512 MB RAM.

| Network | Hugin (sec) | Hugin (MB) | SS (MB) | VE (MB) | RC (sec) | RC Cache (MB) |
|---|---|---|---|---|---|---|
| barley | 4.987+1.522 | 151.137 | 10.499 | 140.637 | 6.43 | 11.002 |
| diabetes | 3.245+0.661 | 88.147 | 4.731 | 83.417 | 3.044 | 5.334 |
| mildew | 1.732+0.561 | 74.895 | 1.91 | 72.985 | 2.434 | 1.907 |
| water | 0.961+0.320 | 30.221 | 3.778 | 26.443 | 1.402 | 3.871 |
| munin3 | 2.133+0.320 | 27.428 | 3.677 | 23.752 | 1.633 | 5.86 |
| munin4 | 5.898+1.423 | 123.353 | 13.951 | 109.402 | 6.76 | 13.281 |
| link | 9.264+3.325 | 318.502 | 12.036 | 306.466 | 17.936 | 14.54 |
| pigs | 0.391+0.080 | 6.361 | 0.95 | 5.412 | 0.351 | 1.02 |
| alarm | 0.020+0.000 | 0.01 | 0.002 | 0.008 | 0.01 | 0.001 |
| b | 0.070+0.070 | 7.276 | 1.055 | 6.221 | 0.43 | 1.008 |

Suppose now that the available memory is limited and we can only cache a subset of the computations performed by RC. The specific subset that we cache can have a dramatic effect on the algorithm's running time. A key question is then to choose that subset which minimizes the running time, a problem referred to as the *secondary optimization problem*. This problem has been addressed in [1], which also discusses a version of RC that not only computes the probability of evidence **e**, but also posterior marginals over families and, hence, posterior marginals over individual variables.[2]

## 3 MEMORY USAGE IN RC

We start with our first analysis of RC which focuses on its space requirements under full caching. We were prompted to look into this issue after realizing that some networks which could not be solved by VE or JT methods due to space limitations were solved relatively easily by RC *without a need to invoke its time–space tradeoff engine,* i.e., these networks were solved under full caching without hitting the same memory limits. This was mostly true for the genetic linkage networks that we discuss in Section 6, whose severe memory requirements have prompted another line of research into time–space tradeoffs [7; 8].

We start our analysis by pointing out that one can talk about three measures of space requirements for VE and JT algorithms. In particular, given an elimination

order $\pi$ and a corresponding jointree $\tau$, one can define the following space models:

*JT/Hugin [11]:* Requires one table per cluster and another per separator in the jointree $\tau$.

*JT/Shenoy–Shafer [20]:* Requires one table per separator in the jointree $\tau$ (assuming that we will only perform the inward pass, otherwise, two tables per separator).

*VE [4]:* Requires one table for each cluster constructed when eliminating a variable in $\pi$. Note that this is no less than the space needed for the cluster tables of Hugin (which is the one we report) and can be a lot more. This is the measure used in [8].

To get a feel of these space requirements on some real–world networks, we list in Table 1 these requirements for some networks from [2; 9]. It should be obvious from this table that the JT/Shenoy–Shafer model is much more space efficient than JT/Hugin and VE.[3] Note that the space requirements of VE are worse than what is shown in the table since a jointree contains fewer clusters than those created by VE. Note also that JT/Shenoy–Shafer and RC are almost the same as far as space requirements are concerned. We will next provide an analytic explanation for this. Before we do this, we point out that Table 1 shows the time to perform the inward pass on a jointree using Hugin system (*www.hugin.com*), a C++ implementation, and the time needed to run RC using the SAMIAM system (*http://reasoning.cs.ucla.edu/samiam/*), a JAVA implementation. Hugin is about twice as fast on these networks. All comparisons in this table are based on an elimination order/jointree constructed by Hugin, which is then converted to a dtree by SAMIAM.

---

[1]Note, however, that RC does not cache results at leaf nodes, so the size of contexts at leaf nodes are not relevant in practice.

[2]The version of RC in [1] uses a *decomposition graph* (dgraph), which is basically a set of dtrees that share structure.

[3]One can develop a special implementation of VE which requires less memory than standard VE; see [6, Sectiom 10.3].

Table 2: Dead Caches in RC.

| Network | Full (MB) | Useful (MB) |
|---------|-----------|-------------|
| barley | 35.579 | 11.002 |
| diabetes | 13.364 | 5.334 |
| mildew | 4.139 | 1.907 |
| water | 9.159 | 3.871 |
| munin3 | 14.187 | 5.86 |
| *munin4* | 32.641 | 13.281 |
| link | 21.036 | 14.54 |
| pigs | 1.508 | 1.02 |
| alarm | 0.005 | 0.001 |
| b | 5.704 | 1.008 |

Table 3: Space for Genetic Linkage Networks.

| Network | Hugin (MB) | SS (MB) | VE (MB) | RC (MB) |
|---------|-----------|---------|---------|---------|
| EA8 | 3.609 | 0.822 | 2.786 | 2.37 |
| EA9 | 229.786 | 20.777 | 209.009 | 26.25 |
| EA10 | 320.013 | 48.064 | 271.949 | 166.63 |
| EA11 | 1,144.075 | 129.135 | 1,014.94 | 112.10 |
| EB6 | 24,485.118 | 4,136.345 | 20,348.773 | 131.71 |
| EB7 | 17,669.819 | 3,984.558 | 13,685.261 | 420.34 |
| EB8 | 6,078.274 | 301.371 | 5,776.903 | 41.96 |
| EB9 | 277.12 | 22.121 | 254.999 | 43.03 |
| EB10 | 12,571.118 | 2,183.242 | 10,387.876 | 1115.81 |
| EC6 | 213.252 | 17.498 | 195.754 | 5.67 |
| EC7 | 81.856 | 6.639 | 75.216 | 9.70 |

One can analytically show that the JT/Shenoy–Shafer model would require less space than the JT/Hugin and VE models, given that each separator is contained in its neighboring cluster, and given that the number of separators is one less the number of clusters in a jointree. But what explains the correspondence between the space requirements of JT/Shenoy–Shafer and RC?

As was shown in [3], the structure of a dtree and its clusters form a jointree. In fact, it is a *binary jointree* since each cluster will have at most 3 neighbors [14]. Moreover, it was shown in [3] that the contexts of a dtree correspond to the separators of induced jointree. This shows analytically that RC on a dtree $\tau$, and JT/Shenoy–Shafer on the binary jointree induced by $\tau$, should require the same space. But binary jointrees are known to take much more space than non-binary ones, and the jointrees in Table 1 are in fact non-binary. Hence, the expectation is that RC should require more space than JT/Shenoy–Shafer since the dtree that SamIam constructs based on a non–binary jointree is binary and, hence, must require more space. We note, however, that the space requirements for RC in Table 1 are after removing *dead caches*. Specifically, if a dtree node $t$ has context($t$) which is a superset of the context of its parent node, then the cache associated with node $t$ is called dead as its entries will never be looked up [3]. Hence, there is no need to cache at node $t$, therefore, reducing the space requirements of RC. Table 2 provides some data on the significant amount of memory reduced due to removing dead caches. Hence, even though dtrees correspond to binary jointrees, they do not suffer from the space disadvantage of binary jointrees due to the removal of dead caches.[4]

Before we close this section, we point to Table 3 which depicts some networks from genetic linkage analysis to be discussed later. It is these networks that prompted us to look into the space requirements of RC under full caching as their space under VE/JT is so demanding that it requires some time–space tradeoff to be invoked under these formalisms [7; 8]. Yet, RC was able to handle them all under full caching. We note here that the dtrees used in this table are not built based on the corresponding jointree, but directly from a min-fill elimination order as described in [3]. Hence, the space requirements of RC and JT/Shenoy–Shafer did not correspond in this table as they did in Table 1.

---

[4]The notion of dead caches can be applied to jointrees, allowing one to avoid storing some separator tables without affecting the running time. This works, however, only if one is interested in inward-pass propagation: the separators will be needed in case an outward-pass is performed.

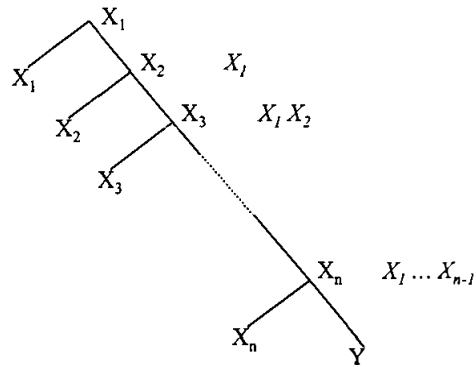

Figure 2: An example dtree.

## 4   REPRESENTING CPTS IN RC

One of the key properties of RC is that it does not require any particular representation of conditional probability tables (CPTs). Specifically, as shown by Algorithm 1, the only function that depends on the CPT representation is LOOKUP, which is given in Algorithm 2. LOOKUP is called on a leaf node $t$ corresponding to network variable $X$ only after all its parents $\mathbf{U}$ have been instantiated to $\mathbf{u}$. If before the call, $X$ is instantiated to $x$, LOOKUP returns $Pr(x|\mathbf{u})$, otherwise, it returns 1. Hence, whether the CPT for $X$ is a table, a decision tree, a set of rules, a formula, or a noisy–or model, that does not affect the statement of the algorithm. As for time complexity of RC, that also remains the same as long as the probability $Pr(x|\mathbf{u})$ can be retrieved in constant time.

This flexibility of RC with respect to the representation of CPTs can have a dramatic effect on its space requirements compared to other methods. Consider for example a simple network where variables $X_1, \ldots, X_n$ are the parents of variable $Y$. Figure 2 depicts a dtree for this network, with the cutset (left) and context (right) shown next to each dtree node. Since the context of each node is a superset of the context of its parent, all caches in this dtree are dead. Hence, RC does not cache any results in this case, requiring no space whatsoever except that needed to store the CPTs. Hence, if the CPT for node $Y$ has a representation which is linear in $n$, then the space complexity of RC on this network is linear in $n$.

More generally, let $N$ be a Bayesian network with a polytree structure with $m$ nodes, where each node has up to $n$ parents. There is a dtree for $N$ under which RC requires $O(m)$ space for caching, and $O(m \exp(n))$ time. Thus, although the worst–case space complexity of RC is exponential in branchwidth, which is $n$ in this case, this is not necessarily the best–case complexity.

We tried to use Netica and Hugin, both of which are based on jointrees, to construct a network with ternary variables $X_1, \ldots, X_{20}$, and binary variable $Y$ which is a noisy–or of its parents $X_1, \ldots, X_{20}$. On a Windows 2000 platform with 1 GB of RAM, neither Netica nor Hugin were able to create this network, both running out of memory as we tried to add the $17^{th}$ parent of $Y$. The reason for this is that these systems will have to convert the noisy–or model to a tabular CPT before they can use the classical jointree method. It is true that there are new variations on the jointree algorithm which attempt to deal with this problem by adopting non–tabular representations, but these extensions require serious algorithmic changes as they have to propose algorithms for replacing classical table operations, such as multiplication and marginalization. The key property of RC is that no algorithmic changes are needed to accommodate non–tabular CPTs, which is a significant advantage that is mostly enabled by the conditioning nature of the algorithm.

## 5 UNIT RESOLUTION

We now turn to the important subject of handling determinism in Bayesian network inference. It has long been observed that the presence of many zeros and ones in the network CPTs can be a source of great savings. One of the earliest approaches to handle determinism is the method of *zero compression* proposed for jointrees [12], which reduces the size of a jointree by eliminating table entries that contain zeros. The method, however, requires that we first build a jointree, and then perform inference on it in order to reduce its size. This method can be extremely effective as long as one can construct the full jointree, which may not be possible for some networks. Another recent method for handling logical constraints was proposed for variable elimination [5; 13] and is closely related to the approach we will propose next for recursive conditioning. We will say more about the relationship between the two approaches later.

The approach we take for handling determinism in RC is based on a key technique in the SAT literature, known as *unit resolution*. Given a logical knowledge base (KB) in the form of propositional clauses, unit resolution is a linear time method for deriving logical implications of the KB based on setting the values of some variables, allowing one to efficiently detect variable assignments which are inconsistent with a KB. Unit resolution is a very important part of any Boolean satisfiability engine, where a major portion of the engine's run time is spent doing it [15]. Our use of unit resolution in Algorithm 1 is for detecting instantiations $\mathbf{c}$ of cutset variables that are guaranteed to have zero probabilities on Line 7, and then skipping these instantiations.

The input to unit resolution is a KB in the form of clauses, where each clause is a disjunction of literals. Each literal is either positive ($X = true$) or negative ($\neg (X = true)$), assuming that every variable has only one of two values. Since Bayesian networks usually contain multi–valued variables, we extend the notion of a literal to either a positive literal ($X = x$) or a negative literal ($X \neq x$).

A clause is satisfied whenever at least one of its literals is satisfied. The basic concept of unit resolution is that when all but one literal $l$ in a clause have been falsified, then literal $l$ must be satisfied in order to satisfy the clause. A literal is falsified when the current setting of variables contradict the literal. A positive literal $X = x$ is satisfied by setting the value of $X$ to $x$. A negative literal $X \neq x$ is satisfied by eliminating the value $x$ from the domain of $X$.

The first step in utilizing unit resolution by RC is to create a KB from a Bayesian network based on the zero/one CPT entries. This can be achieved by making each variable in the network a variable in the KB, and then iterating through each CPT entry to create clauses as follows. Consider the following CPT, where variables $A$ and $B$ have values $\{1, 2\}$ and variable $C$ has values $\{1, 2, 3\}$:

| A | B | C | $Pr(C|A,B)$ | Clauses |
|---|---|---|---|---|
| 1 | 1 | 1 | 1 | $(C = 1 \vee A \neq 1 \vee B \neq 1)$ |
| 1 | 1 | 2 | 0 | |
| 1 | 1 | 3 | 0 | |
| 1 | 2 | 1 | 0 | |
| 1 | 2 | 2 | 1 | $(C = 2 \vee A \neq 1 \vee B \neq 2)$ |
| 1 | 2 | 3 | 0 | |
| 2 | 1 | 1 | .2 | |
| 2 | 1 | 2 | .8 | |
| 2 | 1 | 3 | 0 | $(C \neq 3 \vee A \neq 2 \vee B \neq 1)$ |
| 2 | 2 | 1 | .7 | |
| 2 | 2 | 2 | .3 | |
| 2 | 2 | 3 | 0 | $(C \neq 3 \vee A \neq 2 \vee B \neq 2)$ |

To generate the clauses in the knowledge base, we begin by iterating through the parent instantiations of the CPT for variable $C$. Whenever one of the states $c$ of $C$ has a probability of 1 we generate a clause which contains the positive literal $C = c$, and negative literals $A \neq a, B \neq b$ where $A = a, B = b$ is the corresponding parent instantiation. The first two clauses in the above table are generated using this method. Whenever no state $c$ contains a probability of 1, each state which has a probability of 0 will generate a clause. This clause contains the negative literal $C \neq c$, as well as the negative literals $A \neq a, B \neq b$ where $A = a, B = b$ is the corresponding parent instantiation. The last two clauses in the above table are examples of this.

We also need the following operations on the KB to support unit resolution: assert$(X = x)$, assert$(X \neq x)$, and retract$(X)$ to remove an assertion and any assertions it leads to based on unit resolution. Whenever the knowledge base receives an assertion or retraction, it updates the affected clauses. Anytime all but one literal in a clause is determined to be falsified, it forces the satisfaction of the last literal by calling assert recursively.

The changes needed in RC are then as follows. The operation of "record instantiation **c**" on Line 6 of Algorithm 1 is changed so that it asserts instantiation **c**, and the operation of "un-record instantiation **c**" on Line 8 is changed to retract the instantiation **c**. Moreover, Line 7 will be skipped in case the assertion on Line 6 leads to a contradiction due to unit resolution as this implies a case **c** which has a probability of zero.

We note here that the use of unit resolution has been explored in [5]—and a more general form of constraint propagation has also been explored in [13]—in the context of a variable elimination algorithm, where a KB is also constructed based on the 0/1 probabilities in a Bayesian network. A key difference between these previous approaches and ours is that not only do we exploit unit resolution with respect to the original KB, but also with respect to augmentations of the KB which result from adding further assumptions during conditioning. This provides many more opportunities to discover inconsistencies, which may not exist in the original KB.

We study the effectiveness of our proposed technique for handling determinism in the following section.

# 6  EXPERIMENTAL RESULTS

We will use networks from two different repositories to display the affect of using unit resolution in conjunction with RC, and to provide further data on the actual memory requirements of RC under full caching.

We start with networks from the field of genetic linkage analysis [16], where we use a set of benchmark networks from [7]. The problem here is to compute the probability of some evidence with respect to a given network, known as the *likelihood of the pedigree*. We will not discuss the details of these networks, except to say that (a) they are quite large, containing up to about 9000 variables; (b) require significant memory, upwards of 20 GB for VE and JT/Hugin on some networks; and (3) contain a significant amount of determinism. These challenges make the benchmarks especially suitable for evaluating RC across the dimensions discussed earlier. In fact, the memory demands of these networks have prompted another line of research on time–space tradeoffs, which combines variable elimination and conditioning [7; 8]. Yet, as we shall see next, the memory demands of RC under full caching are modest enough to allow the handling of these networks without invoking its time–space tradeoff engine.

Table 4 depicts the results of running RC on the genetic linkage networks from [7], in addition to reporting the performance of SUPERLINK 1.0 [7] and SUPERLINK 1.1 [8] on these networks. We note here that SUPERLINK is currently the most efficient software system for genetic linkage analysis, and is dedicated to this problem, although it is based on probabilistic inference using a combination of variable elimination and conditioning. Hence, gauging our results with respect to SUPERLINK is quite revealing.

A number of observations are in order about the results in Table 4. First, the results reported for RC and SUPERLINK 1.1 are on a Linux system, with a 2.4 GHz Xeon processor, and 1.5 GB of RAM, while the results for SUPERLINK 1.0 are for a slower machine with 2 GB of RAM and are adapted from [7] (we could not obtain a version that runs on the above platform). Second, given this memory constraint, many of the networks in Table 4 are beyond the limits of standard methods based on VE and JT/Hugin and, sometimes, JT/Shenoy–Shafer; see Table 3. Third, the RC results are based on SAMIAM, using a special imple-mentation which works in log–space given the small numbers involved in genetic linkage analysis. Finally, all dtrees used in these experiments were created from elimination orders based on the min–fill heuristic [3].

Table 4: Genetic Linkage Analysis: Experimental Results.

| File | | Number Variables | RC Time (sec) | Superlink Time (sec) 1.0 | 1.1 | Dtree Cache Size (MB) | KB Size (clauses:literals) | Actual RC Calls | Ratio RC Calls |
|---|---|---|---|---|---|---|---|---|---|
| EA5 | noKB | | 0.2 | | | | | 116.233 | |
| | KB | 2092 | 0.2 | 1.2 | 0.2 | 0.10 | 1588 : 4500 | 70,175 | 1.66 |
| EA6 | noKB | | 0.5 | | | | | 293,507 | |
| | KB | 2491 | 0.4 | 4.7 | 1.6 | 0.28 | 1838 : 5112 | 113,699 | 2.58 |
| EA7 | noKB | | 1.0 | | | | | 651,151 | |
| | KB | 2928 | 0.4 | 3.0 | 0.6 | 0.66 | 2114 : 5812 | 94,583 | 6.88 |
| EA8 | noKB | | 3.1 | | | | | 1,944,307 | |
| | KB | 3790 | 1.3 | 21.0 | 3.3 | 2.37 | 2410 : 6424 | 245,429 | 7.92 |
| EA9 | noKB | | 175.4 | | | | | 64,926,609 | |
| | KB | 7747 | 23.8 | 8510.2 | 356.4 | 26.25 | 4600 : 11784 | 3,347,807 | 19.39 |
| EA10 | noKB | | 685.9 | | | | | 325,061,629 | |
| | KB | 7970 | 201.1 | 10446.3 | 451.7 | 166.63 | 4762 : 12200 | 17,258,401 | 18.83 |
| EA11 | noKB | | 710.4 | | | | | 245,669,493 | |
| | KB | 9027 | 195.0 | > 100 hours | 12,332.0 | 112.10 | 5286 : 13424 | 14,358,315 | 17.11 |
| EB0 | noKB | | 0.2 | | | | | 100,773 | |
| | KB | 1926 | 0.2 | 2.6 | 0.1 | 0.11 | 1751 : 5121 | 33,153 | 3.04 |
| EB1 | noKB | | 0.3 | | | | | 157,915 | |
| | KB | 2317 | 0.2 | 2.6 | 0.2 | 0.18 | 2025 : 5723 | 47,499 | 3.32 |
| EB2 | noKB | | 0.6 | | | | | 396,027 | |
| | KB | 3919 | 0.2 | 82.6 | 0.7 | 0.39 | 2961 : 7831 | 83,307 | 4.75 |
| EB3 | noKB | | 48.3 | | | | | 28,593,739 | |
| | KB | 4710 | 11.6 | 437.6 | 4.1 | 10.00 | 3537 : 9233 | 2,102,409 | 13.60 |
| EB4 | noKB | | 19.1 | | | | | 7,164,295 | |
| | KB | 5088 | 13.0 | 17.3 | 5.5 | 5.80 | 3763 : 9785 | 1,188,043 | 6.03 |
| EB5 | noKB | | 71.7 | | | | | 29,839,491 | |
| | KB | 5483 | 8.0 | 278.8 | 12.4 | 23.82 | 4007 : 10297 | 1,574,619 | 18.95 |
| EB6 | noKB | | 981.1 | | | | | 477,764,683 | |
| | KB | 5860 | 281.9 | 935.9 | 15.5 | 131.72 | 4220 : 10787 | 43,572,385 | 10.96 |
| EB7 | noKB | | 3107.8 | | | | | 1,379,395,611 | |
| | KB | 6240 | 610.3 | 902.8 | 4.0 | 420.34 | 4564 : 11595 | 83,540,347 | 16.51 |
| EB8 | noKB | | 1009.1 | | | | | 322,142,481 | |
| | KB | 6623 | 54.4 | 288.2 | 18.0 | 41.96 | 4945 : 12577 | 4,474,309 | 72.00 |
| EB9 | noKB | | 1656.2 | | | | | 713,453,263 | |
| | KB | 7039 | 123.7 | 114.0 | 27.9 | 43.03 | 5126 : 13011 | 18,814,149 | 37.92 |
| EB10 | noKB | | > 3600.0 | | | | | 1,452,529,921 | |
| | KB | 7428 | 650.4 | 2901.3 | 85.1 | 1115.81 | 5473 : 13994 | 76,309,713 | 19.03 |
| EC5 | noKB | | 26.7 | | | 5.89 | | 8,612,171 | |
| | KB | 1194 | 5.6 | 44.1 | 1.6 | 5.89 | 788 : 2192 | 577,315 | 14.92 |
| EC6 | noKB | | 29.8 | | | 5.67 | | 11,896,861 | |
| | KB | 1200 | 6.3 | 35.3 | 124.9 | 5.67 | 704 : 1968 | 1,298,445 | 9.16 |
| EC7 | noKB | | 59.7 | | | 9.70 | | 19,053,079 | |
| | KB | 1321 | 25.0 | 102.6 | 123.0 | 9.70 | 762 : 2108 | 4,742,827 | 4.02 |

Note that each row in Table 4 contains two lines, one for standard RC (noKB) and another for RC with unit resolution (KB). The running time, actual number of recursive calls, and the ratio of recursive calls, noKB/KB, is also reported. The running time for RC is for two likelihood calculations, while the other values are for a single calculation. Some computations could be stored from one calculation to the next [7], however our current implementation does not use this speedup. Based on this table, it is obvious that the use of unit resolution has only improved performance, considerably in many cases, with some minor exceptions where standard RC was so fast that the use of unit resolution was not justified. To highlight some improvements, consider EB10 which could not be solved within 1 hour, but was then solved in under 10 min-utes using unit resolution. For EB9, the running time dropped from 1656 to 124 seconds. The last column in Table 4 reports on the factor of reduction in the number of recursive calls made by RC, showing a factor of 72 for EB8.

Another key point to observe is the size of KBs generated for these networks, which contained up to about 5000 clauses. This is a relatively large number of clauses, at least by SAT standards, yet is not as large as some of the other networks we shall consider later.

The final set of remarks relating to Table 4 are with regards to gauging the running time of RC against SU-PERLINK, the most efficient software system for genetic linkage analysis. We note here that RC running times only include two likelihood calculations, while SUPERLINK running times additionally include the time to preprocess and create an ordering, as this is the number reported by SUPERLINK. The preprocessing phase used by RC reads in the genetic files, converts it to a Bayesian network, and then simplifies it as discussed in [7].[5] This phase is currently using unoptimized Java code, and along with the dtree generation, took 6 seconds on EA7, 20 seconds on EB6, and 51 seconds on EA11. It should be noted that this only needs to be done once for any number of likelihood calculations, which geneticists would be interested in running. Even after taking this into consideration, RC's numbers appear competitive with the newer version of SUPERLINK on this dataset (see, for example, EA9-11, EB5, EC6-7).

Table 5: KB Compared with No KB.

| Network | RC (KB) | | RC (No KB) | | KB Size | | RC Cache | Ratio |
|---|---|---|---|---|---|---|---|---|
| | Seconds | RC Calls | Seconds | RC Calls | Clauses | Literals | (MB) | RC Calls |
| munin1 | 65.875 | 404,101 | 351.896 | 1,059,450,113 | 8334 | 26500 | 452.9 | 2621.75 |
| munin3 | 2.544 | 201,559 | 3.865 | 24,743,617 | 31826 | 94198 | 8.4 | 122.76 |
| munin4 | 9.293 | 986,675 | 16.844 | 101,264,503 | 39485 | 122441 | 33.5 | 102.63 |
| water | 1.292 | 31,097 | 1.422 | 8,751,957 | 6362 | 35249 | 2.5 | 281.44 |
| munin2 | 14.270 | 582,961 | 2.383 | 14,596,471 | 31365 | 93775 | 4.6 | 25.04 |
| pigs | 2.844 | 1,049,061 | 0.431 | 1,717,165 | 2368 | 7104 | 1.4 | 1.64 |

We note here that SUPERLINK 1.1 is quite recent (the paper to appear). It is based on a new variable ordering technique, which depends on randomization, and appears to be extremely effective [8]. This improvement is orthogonal to the improvement shown by using unit resolution, in that by finding better orderings, we could also generate better dtrees, therefore lifting the additional benefits of SUPERLINK 1.1 to RC. We are currently investigating this direction. We finally note that the benchmarks in [7] included additional networks that we omitted here, as they were too easy, requiring less than a second each.

We now consider another set of networks from a different suite [2], where we will only run one probability calculation. These networks contain even more determinism than the genetic linkage networks, to the point where unit resolution stops being helpful in some cases due to the significant size of KBs. Specifically, consider Table 5 and the munin2 network, which generates a KB with 31,365 clauses. RC with unit resolution reduced the number of recursive calls by a factor of $\approx 25$, yet it slowed down the overall running time by a factor of $\approx 6$. A slowdown is also observed for pigs since the initial running time was very small to start with. An impressive case is the one concerning munin1, where unit resolution reduced the number of recursive calls by a factor of $\approx 2,600$, and reduced the running time from $\approx 352$ to 66 seconds. One key observation about these networks is the significant size of KBs, which explains why the reduction in running time is not proportional to the reduction in number of recursive calls.

We note here that our implementation of unit resolution is quite straightforward, especially that it contains multi-valued variables. There is very little published literature on optimization techniques for unit resolution with multi-valued variables (as opposed to binary variables) and we have yet to invest in such techniques. The promising experimental results we report in this section, however, appear to justify such an investment.

## 7   CONCLUSIONS

This paper rests on two key contributions. First, an analytic and experimental study of the space requirements of RC under full caching, suggesting that it is one of the most space–efficient algorithms for exact inference in Bayesian networks, aside from its ability to allow a smooth tradeoff between time and space. Second, a principled approach for handling determinism in probabilistic inference, which builds upon progress made in the SAT community. Both contributions are supported by experimental results on a variety of real-world networks, including some very challenging networks from the field of genetic linkage analysis. The combination of unit resolution with RC in particular shows a lot of promise in terms of time complexity, suggesting some new directions of research on the interplay between probabilistic and symbolic reasoning.

## Acknowledgements

This work has been partially supported by NSF grant IIS-9988543 and MURI grant N00014-00-1-0617.

---

[5]This involves value and allele exclusion, variable trimming, and merging variables.